\documentclass[conference]{IEEEtran}
\IEEEoverridecommandlockouts
\usepackage{cite}
\usepackage{url}
\usepackage{amsmath,amssymb,amsfonts}
\usepackage{algorithm}
\usepackage{algpseudocode}
\usepackage{graphicx}
\usepackage{caption}
\usepackage{subcaption}
\usepackage{textcomp}
\usepackage{array}
\usepackage[dvipsnames]{xcolor}

\def\BibTeX{{\rm B\kern-.05em{\sc i\kern-.025em b}\kern-.08em
    T\kern-.1667em\lower.7ex\hbox{E}\kern-.125emX}}
\begin{document}

\title{Designing Robust Cyber-Defense Agents with Evolving Behavior Trees}

\author{\IEEEauthorblockN{Nicholas Potteiger, Ankita Samaddar, Hunter Bergstrom and Xenofon Koutsoukos}
\IEEEauthorblockA{\textit{Department of Computer Science} \\
\textit{Vanderbilt University}\\
Nashville, TN, USA \\
\{nicholas.potteiger, ankita.samaddar, hunter.m.bergstrom, xenofon.koutsoukos\}@vanderbilt.edu}}

\maketitle

\begin{abstract}
    Modern network defense can benefit from the use of autonomous systems, offloading tedious and time-consuming work to agents with standard and learning-enabled components. These agents, operating on critical network infrastructure, need to be robust and trustworthy to ensure defense against adaptive cyber-attackers and, simultaneously, provide explanations for their actions and network activity. However, learning-enabled components typically use models, such as deep neural networks, that are not transparent in their high-level decision-making leading to assurance challenges. Additionally, cyber-defense agents must execute complex long-term defense tasks in a reactive manner that involve coordination of multiple interdependent subtasks. Behavior trees are known to be successful in modelling interpretable, reactive, and modular agent policies with learning-enabled components. In this paper, we develop an approach to design autonomous cyber defense agents using behavior trees with learning-enabled components, which we refer to as Evolving Behavior Trees (EBTs). We learn the structure of an EBT with a novel abstract cyber environment and optimize learning-enabled components for deployment. The learning-enabled components are optimized for adapting to various cyber-attacks and deploying security mechanisms. The learned EBT structure is evaluated in a simulated cyber environment, where it effectively mitigates threats and enhances network visibility. For deployment, we develop a software architecture for evaluating EBT-based agents in computer network defense scenarios. Our results demonstrate that the EBT-based agent is robust to adaptive cyber-attacks and provides high-level explanations for interpreting its decisions and actions.  

\end{abstract}


\begin{figure*}
    \centering
    \includegraphics[width=0.7\linewidth]{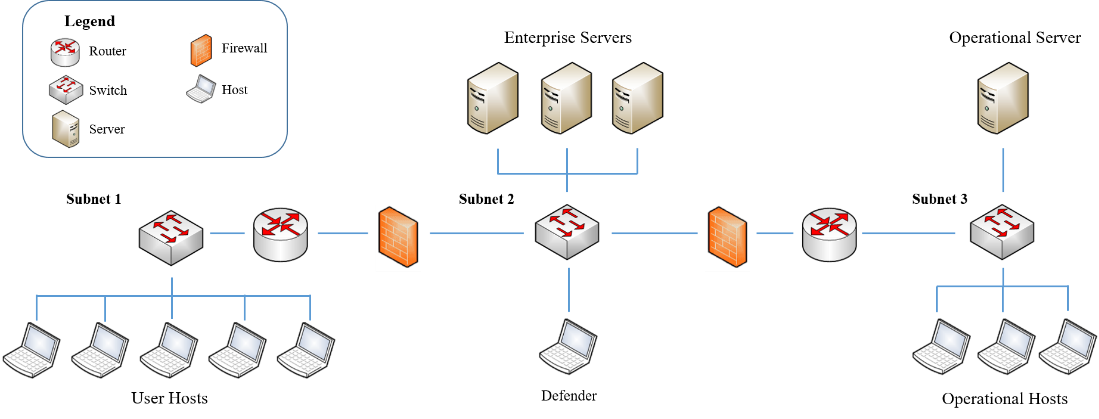}
    \caption{CybORG CAGE Challenge Scenario 2 \cite{cage_challenge_2}; Subnet 1 has five user hosts; Subnet 2 has three enterprise servers and the cyber-defender; Subnet 3 has three operational hosts and one critical operational server.}
    \label{fig:cage_network}
\end{figure*}
\section{Introduction}
\noindent
Modern network defense is an increasingly difficult task due to a multitude of novel and diverse cyber-attacks and the scale of systems. Developers have crafted solutions such as alert monitors and decision logic rules to alleviate human cognitive fatigue, but this is not viable or scalable as new attacks are discovered. The capabilities and strategies of cyber-defense continue to grow and, as such, fully autonomous cyber defense agents are being considered as an alternative to optimally utilize the resources needed to mitigate adversaries. However, there are challenges regarding the transparency of these agents and their robustness. 


Autonomous agents contain a mix of standard and learning-enabled components (LECs) trained with machine learning (ML) and reinforcement learning (RL). These LECs are normally  modeled with neural networks that lack the ability to provide high-level explanations and struggle to complete long-term objectives with multiple subtasks. Cyber-defense breaks down into a variety of subroutines (analysis, monitoring, restoring, deploying decoys, etc.) that a single component will struggle to represent and optimize. 
Additionally, it is unclear how autonomous agents will adapt to multiple varieties of cyber-attacks. Adversaries are versatile, they may use one or more attacks on the system targeting multiple components. 
The adversary may use one strategy at the beginning to explore the system in a low-detection manner and then launch a full-scale attack after they have discovered enough knowledge. The autonomous agent will need to be robust to these scenarios and explain its current perspective of the adversary's behavior. 

An effective way of representing a reactive control policy between subtasks in the hierarchy is using Behavior Trees (BTs) \cite{Colledanchise_2018}. BTs are structures for modeling complex control policies with advantages of modularity, reactivity, and explainability. The reactive nature of the BT allows for explicit switching between multiple behaviors in quick succession depending on environment changes. The LECs and other components can be coordinated and modelled as a policy we refer to as an Evolving Behavior Tree (EBT). An EBT can jointly optimize and model the control flow of optimized components.

The structure of the control of behaviors in an EBT is typically manually developed, but due to the complexity of multiple subtasks and their dependencies, construction can prove tedious and potentially infeasible. Recent work has focused on automatic construction of BTs using genetic programming (GP) \cite{gpbts, gpplannedbts} or large-language models \cite{lykov2023llmbrain, Cao2023RobotBT}. In particular, the works in GP develop abstract environments for computational efficiency that map to a realistic simulation.

In this paper, we develop an autonomous cyber-defense agent that leverages the hierarchical structure of EBTs for robustness against dynamic cyber-attacks. We design the agent in three stages: (1) learning the high-level control structure of the EBT, (2) optimization of LECs for robustness, and (3) integration and deployment to a realistic cyber environment. The research objective of (1) is to enhance scalability by generating control structures without requiring detailed knowledge of network components. This approach enables learning the modular structure of the EBT prior to optimizing LECs, thereby avoiding unnecessary retraining. For (1), we utilize GP and develop a novel abstract cyber-environment referred to as the \textit{Cyber-Firefighter} to map to a realistic cyber-defense simulation scenario and evaluate the structural performance. The research objective of (2) is to develop generalizable behaviors that are robust to uncertainties in a realistic computer network environment. The EBT contains LECs for choosing cyber-agent actions and determining the red agent strategy based on a non-deterministic network state. We develop a software architecture for the construction and integration of an EBT for a computer network defense scenario. In (3), we
deploy and evaluate our autonomous cyber-defense agent in a realistic simulation environment for robustness and explainability.

The main technical contributions of our work are:
\begin{itemize}
    \item Design, optimization, and deployment of an EBT for autonomous cyber-defense of a computer network. The EBT structure is designed using GP with a novel abstract cyber environment, the \textit{Cyber-Firefighter}, that maps to cyber-defense. The EBT structure is generalizable and contains capabilities for decoy deployment, attacker strategy detection, and selection of cyber-operations.
    \item An evaluation of the learned EBT structure in the \textit{Cyber-Firefighter} to demonstrate high-level control performance against an attacker in a network. The GP algorithm in tandem with the \textit{Cyber-Firefighter} is successful at learning an EBT structure that maximizes the performance metric (fitness) to promote mitigation and visibility of an attack.
    \item Development of a software architecture to support the construction and deployment of EBTs on a computer network. The architecture utilizes a blackboard of data sources in a publish-subscribe method to facilitate interaction between the EBT and computer network environment. The computer network environment in this paper is CybORG \cite{cyborg_acd_2021}, an abstracted version of a computer network, compatible with ML and RL algorithms.
    \item An evaluation of the robustness and explainability of the EBT in CybORG \cite{cyborg_acd_2021} using CAGE Challenge Scenario 2 \cite{cage_challenge_2}: a computer network task where the agent must defend against an adversarial agent. We develop an adversarial red agent for this scenario that switches strategies during execution to evaluate the adaptation of our approach. The EBT is successful at defending against dynamic attacks with a $39\%$ increase in the average reward compared to a state-of-the-art method in CybORG CAGE Challenge Scenario 2. 
    The explainable nature of the EBT allows us to monitor key events, such as when the strategy switches or a decoy is deployed, and model transitions between high-level subtasks.  
\end{itemize}



\section{Related Works}
\noindent
With the onset of sophisticated cyber-attacks, traditional security measures often fall short. Thus, there is a need for more advanced and interpretable solutions. Nowadays, different RL methods are adopted to develop learning enabled cyber-defense policies in autonomous networks~\cite{kiely2023autonomous}. Different RL-based cyber-defense policies have used CybORG to enact their approaches. Foley \emph{et al.} utilized a goal-conditioned hierarchical RL (HRL) to select trained defense strategies in CybORG CAGE Challenge Scenario 2~\cite{mindrake}. The defense strategy in~\cite{mindrake} is optimized via RL for each attacker strategy, following which a meta-policy selects between the defense strategies based on the attacker behavior at the beginning of a simulation. The winners of the CAGE Challenge Scenario 2, the \textit{CardiffUni}, used a similar goal-conditioned approach but with an added focus on reducing action space and deploying decoys~\cite{mindrake}. Wolk \emph{et al.} presented an alternative where an ensemble approach aggregates the policy output~\cite{wolk2022cage}. Towards emulation, Molina-Markham \emph{et al.} developed a novel tool, FARLAND, with a focus on realistic cyber-defense environments and curriculum learning for cyber-defense agents~\cite{molinamarkham2021network}. Unlike classic RL tasks where the agents are regularly rewarded for progress, the reward signals assigned to cyber-security tasks are distributed sparsely across each defense episode in the form of penalties. To overcome this gap, Elizabeth \emph{et al.} presented a reward shaping policy in deep RL and evaluated the policy in CybORG~\cite{elizabeth2023}.

Another line of study is motivated by neurosymbolic AI approaches which combine the pattern recognition capabilities of neural networks along with the explicit reasoning of symbolic systems. Neurosymbolic AI serves as an emerging area of research in autonomous cyber-defense~\cite{neurosymbolicAI2023}. An increasingly popular approach for designing neurosymbolic autonomous agents is using Behavior Trees (BTs) \cite{Colledanchise_2018}. BTs are widely used in control architectures, computer games, robotics, etc~\cite{bt2018}. 
RL can be employed to learn complex BT behaviors. It provides the flexibility and ability to discover innovative solutions for sub-tasks within BTs. Whether optimizing a single behavior or multiple learning-enabled behaviors within the tree, RL or HRL techniques are used to jointly optimize and generate policies for BT tasks. Recent efforts have focused on integrating HRL into the BT learning process~\cite{starcraftBT, Li2021MixedDR}. These works utilize hierarchical option policies~\cite{bacon2017option} to coordinate a finite set of unknown subtasks in video game simulation environments. Our prior work applied goal-conditioned HRL and a manually constructed EBT for a maze navigation task \cite{potteigerebts}. In this work, we develop an approach that combines the EBTs with LECs to provide autonomous cyber-defense in computer networks. To the best of our knowledge, this is the first work in the literature that models autonomous cyber-defense agents using EBTs to analyze system behavior and apply appropriate mitigation tactics against adversaries.

\section{Autonomous Cyber-Defense}
\noindent
Autonomous cyber-defense is the long-term task of fortifying and mitigating a system against cyber-attackers without human intervention. It is a task that constantly needs to be iterated on as long as the system is operating, otherwise novel attacks could breach and disrupt critical resources. Cyber-defense agents can utilize a mix of security standards and LECs to successfully adapt and defend a system.


LECs typically rely on function approximators, such as neural networks, trained with ML or RL to compute optimal actions. Existing algorithms work well for short-term tasks. However, as the task increases in complexity and longevity, new subtasks and capabilities are required that a single component struggles to optimize. Also, the components are not transparent, leading to challenges with assurance and trustworthiness. In a system with safety-critical resources, it is essential that we understand the behavior of an agent, or else there can be unintended consequences due to agent error.

In general, autonomous cyber-defenders need to be prepared for a diversity of potential cyber-attacks. These attacks can derive from one or multiple actors. During an attack, an actor may decide that it should switch to another attack strategy based on information it has observed in the system. If the cyber-defender is not aware of this switch, the system may be vulnerable due to a lack of knowledge of defense against the new strategy.

In the face of these challenges, the objective of this paper is to develop a neurosymbolic model representation of an autonomous cyber-defense agent that 
(1) captures an explicit hierarchy of subtasks and the control flow required to execute subtasks,
(2) employs LECs for specific subtasks, and
(3) adapts to multiple dynamic cyber-attacks. The model must allow optimization of the LECs and generalize to multiple attack scenarios. Furthermore, after training, the model must be deployable to a system for evaluation.

We consider CybORG \cite{cyborg_acd_2021}, a complex network simulation environment that abstracts real-world scenarios, as our canonical problem for developing our model. The network scenario we consider is CAGE Challenge using Scenario 2 \cite{cage_challenge_2} as shown in Fig.~\ref{fig:cage_network}. The network consists of three subnets: Subnet 1 consists of user hosts that are non critical, Subnet 2 consists of enterprise servers to support user activities on Subnet 1 and Subnet 3 consists of critical operational server and three user hosts.
CybORG provides an interface to construct attacker (red) and defender (blue) agents with learning-enabled capabilities in a seamless manner. During simulation, the red and blue agents execute actions in parallel. The red agents can scan hosts and subnets, launch exploits, raise privileges, or disrupt a compromised host. To mitigate red agent behavior, the blue agents can take no action (\textit{Sleep}), monitor the network (\textit{Monitor}), analyze a host (\textit{Analyze}), remove any malicious software from a host (\textit{Remove}), restore a host back to a good state and remove the attacker (\textit{Restore}), or deploy one of seven decoy service types on a host (\textit{Deploy Decoy}).

Different red agent strategies can be constructed using CybORG. CAGE Challenge Scenario 2 uses two types of pre-defined red agent strategies. The first is \textit{BLine}, which uses full knowledge of the network to traverse towards the critical operational server. The second strategy is \textit{Meander} that seeks to explore and disrupt the network, compromising each host before moving to the next one. 

We consider a third strategy \textit{RedSwitch} that combines \textit{BLine} and \textit{Meander}. This strategy first instantiates an agent using the \textit{Meander} strategy, then after a randomly selected amount of time, the attacker transitions to an agent using the \textit{BLine} strategy. The intuition behind this red agent switching strategy is that an attacker may first explore the network (\textit{Meander}) to discover its topology before deploying a more disruptive attacker (\textit{BLine}).

\begin{figure}
    \centering
    \includegraphics[width=\linewidth]{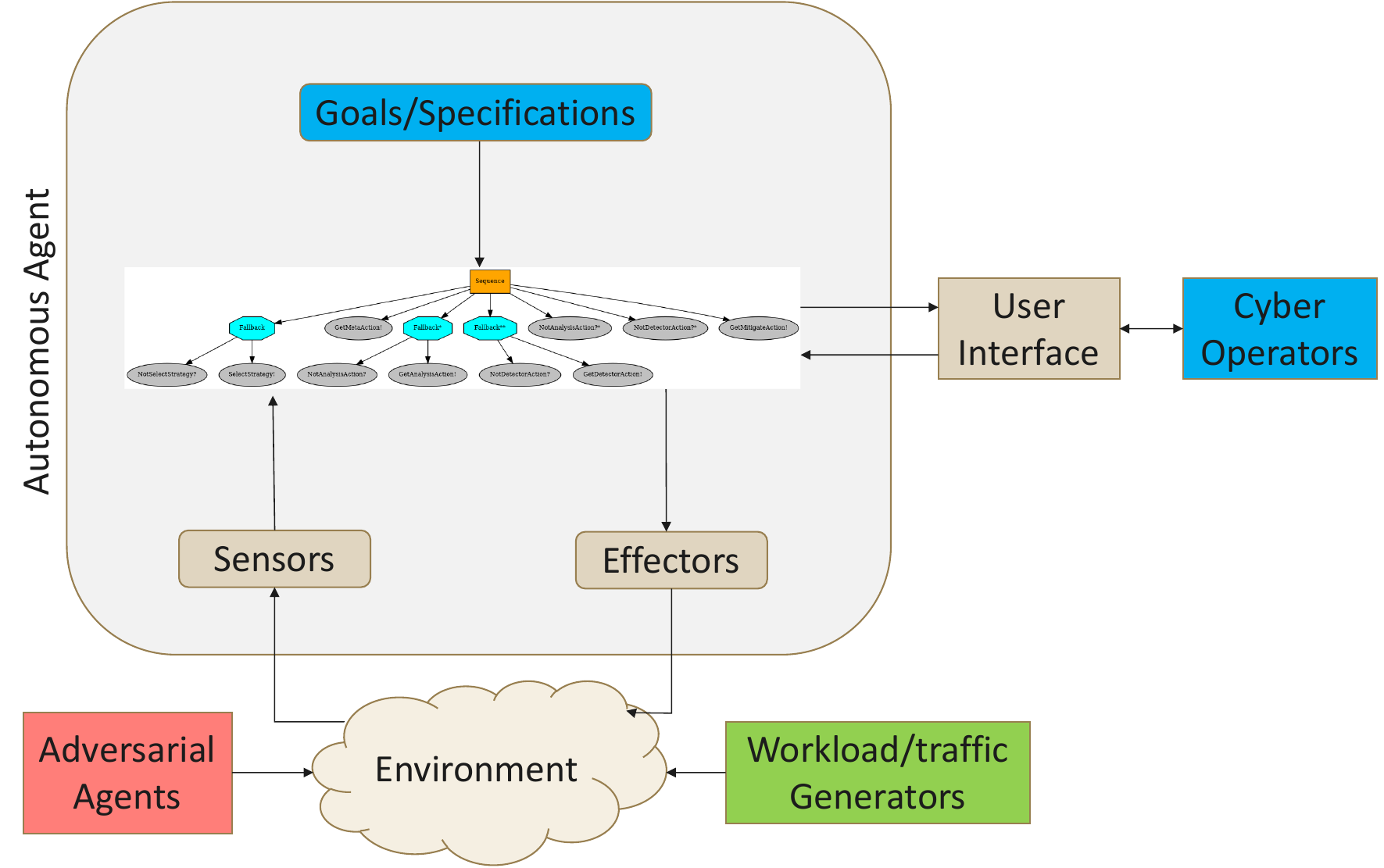}
    \caption{Autonomous Cyber-Defense Agent}
    \label{fig:defense_agent}
\end{figure}

\section{Optimal BT Structure for Cyber-Defense}
\noindent
We present a neurosymbolic approach to autonomous cyber-defense that is robust against dynamic cyber-attacks. The agent constructed interacts with the environment using cyber-agent actions from a set of capabilities to mitigate adversarial red agents. Fig.~\ref{fig:defense_agent} presents the control flow of the autonomous cyber-defense agent. Given a goal or specification, a symbolic structure acts as a model to interact with the environment. The symbolic structure utilizes cyber actions and sensor information to take effective action(s) via effectors against adversarial agents in the environment.


The symbolic structure in our autonomous cyber-defense agent is a BT. The BT allows us to reason about cyber-defense control at a high level and provides reactive switching to adapt based on environmental shifts. Additionally, BTs are modular, allowing new capabilities to be seamlessly integrated. 

\begin{figure*}
    \centering
    \includegraphics[width=0.5\linewidth]{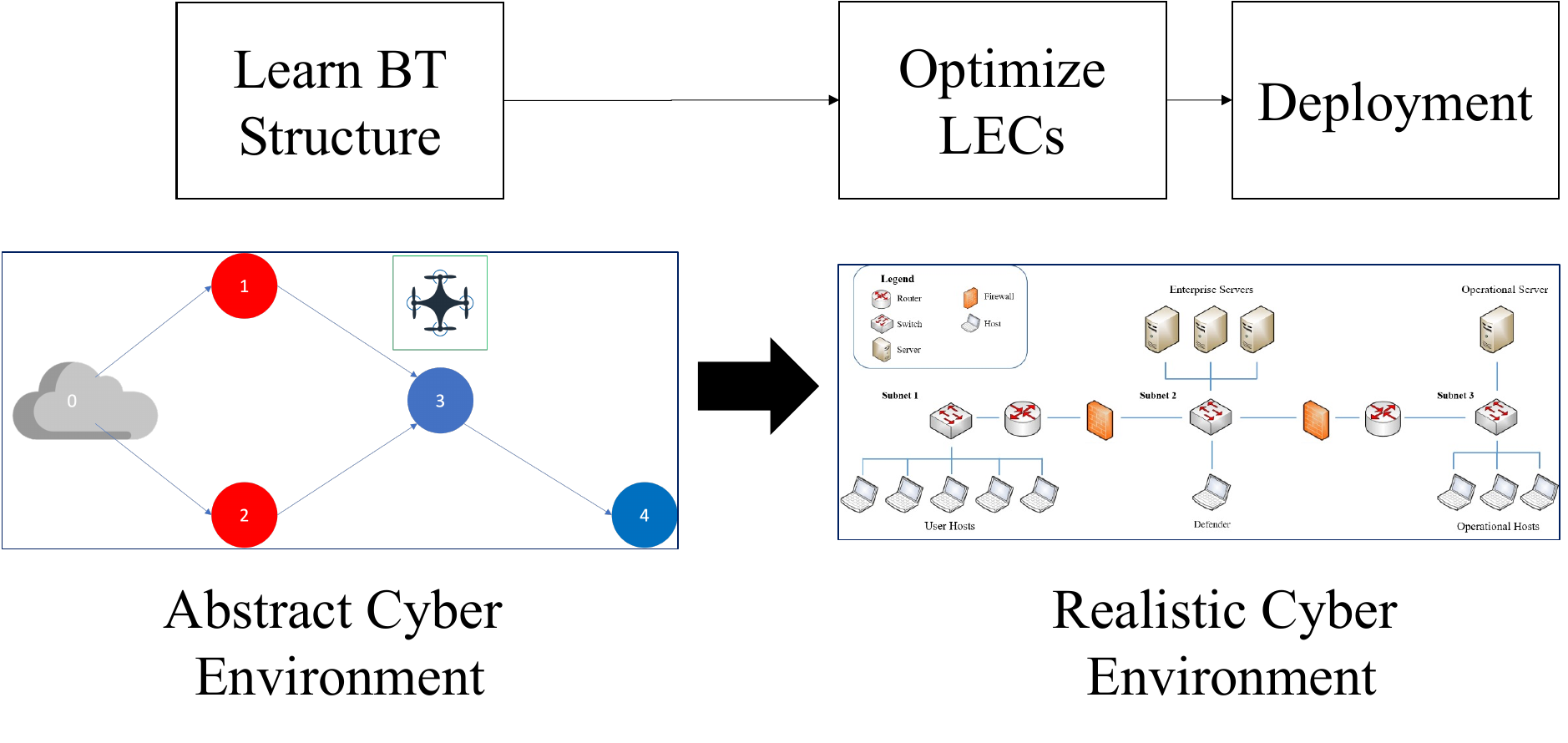}
    \caption{Robust Autonomous Cyber-Defense EBT Design Approach.}
    \label{fig:framework}
\end{figure*}

In this section, we describe the optimization method utilizing GP to compute a BT structure for cyber-defense control. The first stage of the neurosymbolic design approach, in Fig.~\ref{fig:framework} focuses on learning the structure and high-level control of behaviors in the BT. A key contribution of learning the structure is the development of a novel abstract cyber environment, \textit{Cyber-Firefighter}, to capture high-level computer network defense. Another benefit of the abstract environment is efficient simulation for evaluation of the structure of a BT over a more computationally heavy realistic simulator.

\subsection{Cyber-Firefighter Abstract Environment}
We design an abstract environment, \textit{Cyber-Firefighter}, a partially-observable pursuit-evasion game, where a defender must contain an attacker from spreading over a network. The pursuit-evasion game is inspired by the \textit{Firefighter} \cite{bonato2022invitation}, a game where a ``firefighter" (defender) must optimally contain a spreading ``fire" (attacker) in a network of ``trees" (nodes) using full knowledge of the ``fire" movements (attacker's movements) and a ``fire retardant" (mitigation action) is executed to block the spread. The game terminates when the ``fire" can no longer spread to new ``trees".

In the \textit{Cyber-Firefighter} game, the objective remains the same. However, to make the game similar to cyber-defense on a computer network, the ``firefighter" does not have full knowledge of the ``fire" movements (attacker's movements). Therefore, we re-frame the problem as a partially observable environment and include two new actions to reveal information in the environment. 1) The ``firefighter" can deploy ``drones" (detectors) to detect the ``fire" if it spreads into a ``drone" detection zone. 2) The ``firefighter" can activate a ``drone" (perform analysis) to increase the detection zone and gain further information from the environment. Once the information, \{the ``fire" has spread to\} is revealed to the ``trees" (nodes), the ``fire retardant" (mitigation action) is placed (executed) to block the spread. 

Formally, the \textit{Cyber-Firefighter} environment can be represented as a network graph, $G: (V, E)$, where $V$ denotes the set of ``trees" or nodes and $E$ denotes the set of edges or connections through which the ``fire" can spread. The neighbor of a node $v$ is denoted by $N(v) = \{u\ |\ (v, u) \in E;~v, u \in V\}$. $N_r(v)$ denotes the nodes of the induced neighbor subgraph up to a radius $r$ from node $v$. 

Table~\ref{table:notation} shows the notations to describe the \textit{Cyber-Firefighter} environment.
\begin{table}[h!]
\centering
\begin{tabular}{| >{\vspace{3pt}}m{2.3cm} | >{\vspace{3pt}}m{5.7cm} |}
\hline
\textbf{Notation} & \textbf{Description} \\
\hline
$v_f \in V$ & fire source \\
\hline
$T_{burn}^{max} \in \mathbb{N}$ & maximum burning time of any node \\
\hline
$T_{burn}^t(v) \in \mathbb{N}$ & burning time of node $v$ at timestep $t$\\
\hline
$T$     & total number of timesteps upto which the game is played\\ 
\hline
$r \in \mathbb{N}$ & drone search radius \\
\hline
$R_t \subset V$ & set of retardant nodesa at timestep $t$, \emph{i.e.}, set of nodes that contain the ``fire" spread using flame retardant\\
\hline
$I_t \subset V$ & set of visibility nodes at timestep $t$, \emph{i.e.}, set of nodes that are visible from the deployed drones\\
\hline 
$D_t(v) \in \{-1, 0, 1\}$ & status of a drone deployed on node $v$ at timestep $t$, ($-1 :$ no drone deployed on node $v$, $0 :$ drone is deployed but not activated, $1 :$ drone action is activated)
\\
\hline
\end{tabular}
\caption{Notation and Descriptions}
\label{table:notation}
\end{table}

The state $s$, where $s\in \mathcal{S}$ and  $\mathcal{S}$ is the state space of the \textit{Cyber-Firefighter} game, contains a snapshot of the burning time of visible nodes $I_t$, retardant nodes $R_t$, the status of the drones deployed on all the nodes in the network, and the network graph $G$. At each timestep $t$, an action $a_t \in \mathcal{A}$ can be executed, where $\mathcal{A}$ is the action space. Each action $a_t \in \mathcal{A}$, is a tuple of two variables, $(v_t, type_t)$, where $v_t \in V$ and $type_t$ denotes the type of action taken and can take any value from $\{0 : $ \textit{Deploy Drone}, $1 : $ \textit{Activate Search}, $2 :$ \textit{Place Retardant}$\}$. 


Algorithm~\ref{alg:cyber_fire} presents the \textit{Cyber-Firefighter} game execution with graph $G$, fire source $v_f$, drone search radius $r$, state space $\mathcal{S}$,  action space $\mathcal{A}$ and a ``firefighter" action policy $\pi:\mathcal{S} \to \mathcal{A}$ as inputs. At each timestep $t$, the game applies policy $\pi$ on state $s_t$ and gets an action as output (line~\ref{alg:line:action}). 
The action type $type_t$ is executed for node $v_t$, updating visible $I$, drone status $D(v_t)$, and retardant $R$ nodes in lines~\ref{alg:line:start_types}-\ref{alg:line:end_types}. Then the burn times $T^t_{burn}(v)$ for all nodes $v\in V$ are updated in lines~\ref{alg:line:fire_update_start}-\ref{alg:line:fire_update_end} and the state is updated to $s_{t+1}$ in line~\ref{alg:line:update_state}. The game terminates 
when the condition in line~\ref{alg:line:termination} is satisfied.

\begin{algorithm}
\caption{Cyber-Firefighter($G : (V, E)$, $v_f$, $r, \mathcal{S}, \mathcal{A}$, $\pi$)} \label{alg:cyber_fire}
\begin{algorithmic}[1]

\State Initialize $I_0$, $R_0$ to $\emptyset$ and $t$ to $0$;
\State Initialize $D_0(v)$ to -1, $\forall$ $v \in V$;
\State Initialize $T_{burn}^0(v_f)$ to $T^{max}_{burn}$;
\State $s_0 = \{T_{burn}^{0}(v),\ \forall v\in I_0, R_0, D_0(v)\ \forall v \in V, G\}$\label{alg:line:init_state}

\While{$t < T$ $||$ ($t > 0$ $\&\&$ $T_{burn}^{t-1}(v) = T_{burn}^{t}(v)\ \forall v \in V$)} \label{alg:line:termination}
    \State $v_t, type_t = \pi(s_t)$\label{alg:line:action};\Comment{Select Action}
    \If{$type_t == 0$} \label{alg:line:start_types} \Comment{Deploy Drone}
        \If{$D_t(v_t) == -1$}
            \State \textbf{Set:} $D_{t+1}(v_t) = 0$, $I_{t+1} = I_t \cup \{v_t\}$;
        \EndIf
    \ElsIf{$type_t == 1$} \Comment{Activate Search}
        \If{$D_t(v_t) == 0$}
            \State \textbf{Set:} $D_t(v_t) = 1$, $I_{t+1} = I_t \cup \{N_r(v_t)\}$;
        \EndIf
    \ElsIf{$type_t == 2$} \Comment{Place Retardant}
        \If{$v_t \in I_t$ $\&\&$ $T_{burn}^t(v_t) \neq T_{burn}^{max}$}
            \State \textbf{Set:} $R_{t+1} = R_t \cup \{v_t\}$;
        \EndIf
    \EndIf \label{alg:line:end_types}
    \For{$u\in V \ \backslash \ R_t$} \Comment{Update Burn Time}\label{alg:line:fire_update_start}
        \For{$v' \in N(u)$}
            \If{$T_{burn}^{t}(v') == T_{burn}^{max}$}
                \State \textbf{Set:} $T_{burn}^{t+1}(u) = \min(T_{burn}^{t}(u) + 1, T_{burn}^{max})$;
                \State \textbf{Break};
            \EndIf
        \EndFor     
    \EndFor\ \label{alg:line:fire_update_end}
    \State $B = T_{burn}^{t+1}(v)\ \forall v\in I_{t+1}$;
    \State $O = D_{t+1}(v)\ \forall v \in V$;
    \State $s_{t+1} = \{B, R_{t+1}, O, G\}$; \Comment{Update State}\label{alg:line:update_state}
\EndWhile
\end{algorithmic}
\end{algorithm}

Given a network graph $G: (V, E)$ and a fire source $v_f$, the objective of the ``firefighter" $\pi$ is to limit the number of ``trees" with the maximum burn time ($T_{burn}^{max}$) over $T$ timesteps using actions in $\mathcal{A}$. The minimization objective function, $M$, can be represented as: 

\begin{equation}
    M \triangleq \min_V |\{ T_{burn}^{[0,T]}(u) = T_{burn}^{max}, \forall u \in V \} |
    \label{eq:target}
\end{equation}

The action policy $\pi$ of the ``firefighter" selects the actions following Equation~\eqref{eq:target} (line~\ref{alg:line:action}).

\subsection{Cyber BT Behaviors}
The construction of a BT requires a set of behaviors as building blocks. The behaviors are generalizable, \emph{i.e.}, they can be mapped to our abstract environment as well as our realistic cyber-security environment, CybORG. There are a variety of behaviors that are used to construct the BTs \cite{Colledanchise_2018}.  In the construction of a BT, each timestep is known as a \textit{tick}. A BT \textit{tick} starts from the root behavior and follows a \textit{Depth-First Traversal}. This traversal can shift depending on the returned status of child behaviors. The behavior(s) can return a status of \textit{Failure}, \textit{Running} or \textit{Success} altering the execution traversal of the BT.
 
Behaviors can be broadly classified into two groups: control behaviors and execution behaviors. Control behaviors are the internal behaviors in a BT that control the logical flow of switching between the behaviors. Execution behaviors are the leaf behaviors in a BT that execute the selected actions in the environment. The control behaviors in the BT can be either \textit{Sequences} or \textit{Fallbacks}. \textit{Sequences} execute a set of child behaviors sequentially until all child behaviors return \textit{Success}, return \textit{Failure} otherwise. \textit{Fallbacks} execute child behaviors until one child behavior returns \textit{Success}, return \textit{Failure} otherwise. The execution behaviors in the BT can be either \textit{Condition} or \textit{Action} behaviors. The return status of the execution behavior is determined based on its intended logical condition or user-defined functionality. After the status is returned, it propagates back up to the root, recursively updating the status of the parent control behaviors.

We define five \textit{Action} behaviors, to allow the BT to defend against an adversary:  
\begin{enumerate}
    \item \textit{SelectStrategy!} selects a defense strategy based on an adversarial movement.
    \item \textit{GetMetaAction!} selects one of three defense behaviors using the selected defense strategy.
    \item \textit{GetDetectorAction!} deploys a detection mechanism in the environment to alert a local adversarial activity.
    \item \textit{GetMitigateAction!} prevents an adversary from achieving their objective, such as blocking adversarial movement or restoring a network node to a previously ``good" state.
    \item \textit{GetAnalysisAction!} monitors or analyzes the environment retrieving new information that is unknown to the agent.
\end{enumerate}
 
Furthermore, there are four condition behaviors. There is one condition behavior for each of the three defense operation behaviors, to ensure a behavior is only enacted when chosen by \textit{GetMetaAction!}. Additionally, there is a condition behavior for \textit{SelectStrategy!} to ensure the strategy is only selected or shifted when necessary. 

To construct BTs for execution in the \textit{Cyber-Firefighter} environment, we map the defined BT behaviors to actionable behaviors in the environment. The three defense behaviors map to the three ``firefighter" actions. Each defense behavior determines a node to deploy a drone (\textit{GetDetectorAction!}), activate a drone for further visibility analysis (\textit{GetAnalysisAction!}), or place fire retardant to block the fire spread (\textit{GetMitigateAction!}. \textit{GetMetaAction!} selects which operation to perform using a strategy. In this paper, the strategy defines a state-action policy for choosing operations based on the current state of the fire and visibility of the network. \textit{SelectStrategy!} determines the appropriate strategy to enact based on the state of the environment. For example, a strategy could be to repeat the process of deploying a drone, activating a drone, or placing a retardant.

\begin{figure*}
    \centering
    \includegraphics[width=\linewidth]{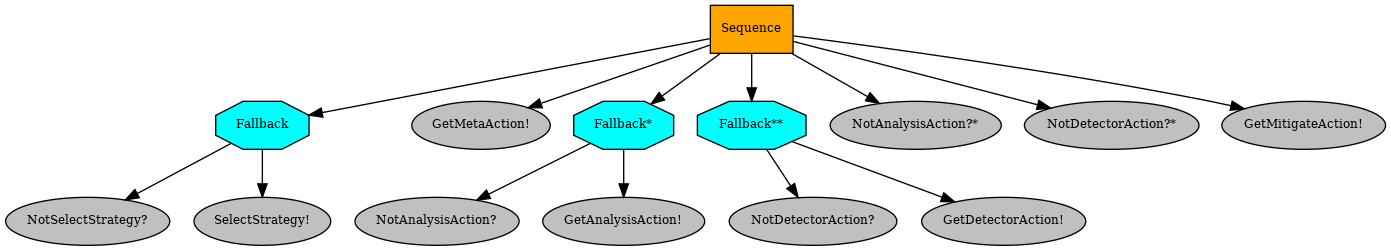}
    \caption{Learned GPBT Architecture for Strategy Switching}
    \label{fig:bt_structure}
\end{figure*}

The behaviors can be arranged through learning or expert knowledge into the BT structure as described in Fig.~\ref{fig:bt_structure}. The root of the BT in Fig.~\ref{fig:bt_structure} is a \textit{Sequence} behavior with $7$ child behaviors. From left-to-right the behaviors are executed. The first behavior is a \textit{Fallback} behavior that determines if a new strategy should be selected given two children \textit{NotSelectStrategy?} and \textit{SelectStrategy!}. If \textit{NotSelectStrategy?} returns a status of \textit{Failure} indicating a strategy should be selected, then the \textit{Fallback} will execute the second behavior \textit{SelectStrategy!}. Then the root \textit{Sequence} will execute the second child \textit{GetMetaAction!} to select a cyber-operation behavior to perform. The next $5$ child behaviors of the root \textit{Sequence} relate to selecting and executing the correct cyber-operation behaviors in a similar manner.


\subsection{Learning BT Structure using Genetic Programming}
To learn the optimal structure of the EBT in Fig.~\ref{fig:bt_structure}, we use GP combined with suboptimal baseline BTs. The inputs to the GP are the pre-defined BT nodes for the \textit{Cyber-Firefighter} environment and a baseline BT that only considers strategy selection. The abstract environment will be used for evaluation and computation of a fitness value for optimizing towards an efficient BT that mitigates the attacker spread on the network.

More specifically, the GP algorithm is initialized with the set of execution behavior nodes and control behavior nodes. The BT is defined and generated randomly from these set of behaviors. The intention of the baseline BT is to provide a starting representation to encourage expansion and enhance training efficiency. After initialization, the GP algorithm selects, breeds, and evaluates BTs for a fixed number of generations using a fitness function, ultimately computing a solution that maximizes the fitness.

The fitness function, $F(x)$, is the primary component in GP that drives the definition of the optimal BT. We consider the ``fire" impact, the visibility of the network, and the size of BT nodes as a fitness function for the \textit{Cyber-Firefighter} environment. We define the fitness function as 

\begin{equation}
    F(x) = c_v |I| - c_l |x| - c_f \sum_{t=0}^{T} | \{v \in V \ | \ T_{burn}^{t}(v) = T_{burn}^{max}\} |
\end{equation}

where $|I|$ is the number of visible nodes in the environment. $|x|$ is the number of BT behaviors in the BT. The ``fire" impact is measured by the number of nodes with the maximum burn amount at each timestep $t$. The coefficients $c_v, c_l, c_f$ are assigned to each term to weigh the impact of each term on overall fitness. Optimizing $F(x)$ maximizes the number of visible nodes and minimizes the number of BT behaviors and the cumulative ``fire" impact over $T$ timesteps. Therefore, optimizing toward the maximum $F(x)$ encourages a compact and efficient BT that can gain visibility in the network and mitigate adversarial activity, by reducing ``fire" spread, using a minimal number of behaviors.

\section{Robust BTs with LECs}
\noindent
Behaviors from the abstract cyber environment must be transferred and implemented for the realistic environment, CybORG CAGE Challenge Scenario 2, as shown in Fig.~\ref{fig:framework}. Behaviors that require long-term objectives, such as cyber-defense against unknown attackers over an extended period, benefit from the integration of optimized LECs for robustness and generalizability. There are two LECs for our approach that focus on decision-making for selecting a cyber action behavior and switching strategies. Optimal LECs are then connected to behaviors in the BT structure for deployment in a new structure that we refer to as an Evolving Behavior Tree (EBT).

In this section, we first describe the transfer of behaviors to the realistic cyber environment, then derive the optimization techniques and policies for the LECs. At the end of the section we integrate the LECs with the BT structure to form the EBT for deployment.

\subsection{Behavior Transfer in Realistic Cyber Environment}
An optimal BT structure learned in an abstract environment must have its behaviors mapped to the realistic cyber environment for optimization and deployment. The realistic cyber environment employed in this paper is the CybORG CAGE Challenge Scenario 2. 

The cyber-defense actions are mapped to their appropriate behaviors. \textit{GetAnalysisAction!} maps to \textit{Analyze} and \textit{Monitor} as they reveal information about the computer network state. \textit{GetDetectorAction!} maps to a greedy deterministic policy that selects a \textit{Deploy Decoy} action conditioned on a host node. Furthermore, \textit{GetMitigateAction!} maps to \textit{Remove} and \textit{Restore} because the actions prevent further damage caused by the attacker. To coordinate which of the above behaviors are selected, a cyber-agent controller policy is used that is executed by \textit{GetMetaAction!} bahavior. Finally, the type of controller policy used is determined using the \textit{SelectStrategy!} behavior. A strategy switching policy captured by \textit{NotSelectStrategy?} behavior determines if the controller policy needs to be switched.

\subsection{Optimizing Behaviors}
Components in the EBT with complex decision-making benefit from optimization techniques for generalizability and adaptation to novel scenarios. Both the cyber-agent controller and the strategy switching will need to be optimized to mitigate complex red agent behavior.

The cyber-agent controller contains a set of $N$ learnable policies, $[\pi_1, \pi_2, ... \pi_N]$, for $N$ red agent strategies. Each policy $\pi_i$ has an associated reward $r_i$ that maximizes with standard RL to mitigate red agent strategy $i$. $\pi_i$ inputs an observation from the network and outputs a specific cyber-agent. In this paper, $r_1 = r_2 = \ldots = r_N$, as we use the reward provided by CAGE Challenge Scenario 2 \cite{cage_challenge_2} which quantifies the disruption to the network caused by red agents. A larger negative value indicates higher disruption.  

During execution, the cyber-agent controller chooses an action from $\pi_i$ that is associated with the current red agent strategy. We predict red agent strategy $i$ using a strategy switching policy $\pi_{strat}$. $\pi_{strat}$ is a temporal policy that inputs a window of observations and outputs a red agent strategy labelled $i$. The policy is trained using supervised learning from a collection of \textit{(observation, strategy label)} pairs. Collecting pairs is achieved through using a fully observable strategy switching policy.

\subsection{Deployment}
The optimized LECs can be connected back to their respective behaviors for deployment. Behaviors with LECs such as \textit{GetMetaAction!} are implemented to allow for appropriate querying and functionality the LECs required to construct its policy. This involves identifying key inputs, such as the state of the system, and outputs, such as the meta-action. A behavior can have input and output dependencies on another behavior due to this setup. Behaviors with standard components are implemented in a similar manner for their full intended functionality. The fully implemented EBT can then be deployed and executed on the intended realistic cyber-environment. This cyber-environment can either be a simulation with abstractions or an emulation with realistic hardware and computer software components.

\section{Evaluation}
\noindent
We perform an evaluation of optimizing the BT structure and LECs of the EBT.
We determine if the optimal structure can appropriately mitigate an adversary and gain visibility of the network. For the EBT we determine if it is robust to multiple cyber-attacks compared to a state-of-the-art solution. We track the performance of simulations of \textit{Cyber-Firefighter} and CybORG CAGE Challenge Scenario 2. We also discuss the explainable nature of the EBT and its role in monitoring the program flow.

\begin{figure*}
    \centering
    \includegraphics[width=0.75\linewidth]{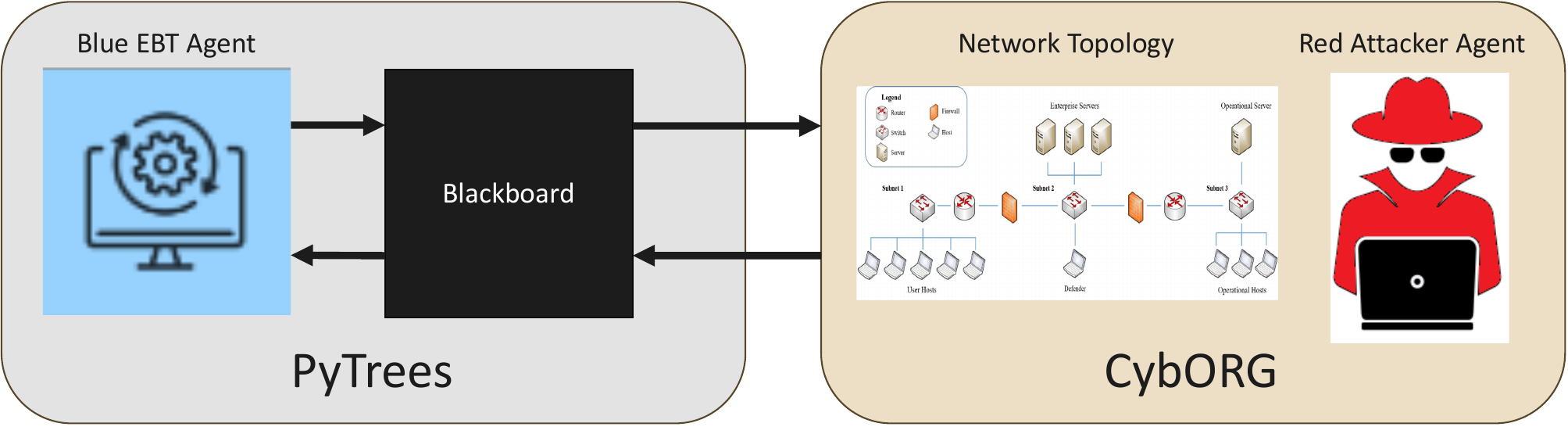}
    \caption{EBT and CybORG Software Architecture. A blackboard is used as a communication mechanism between the EBT and CybORG simulation.}
    \label{fig:software_arch}
\end{figure*}

\subsection{Software Architecture}

For evaluation, we developed a software architecture, shown in Fig.~\ref{fig:software_arch}, that constructs and executes an EBT with a computer network simulator. A communication mechanism known as a \textit{blackboard}~\cite{pytrees} is instantiated between the the EBT and the simulator to maintain and update shared data. Similar to a publish-subscribe framework, the EBT and simulator can read or write particular data values based on access permissions. Behaviors in the EBT are restricted to access only variables on the blackboard specific to their functionality to prevent data leaking. To develop the EBT that are used for high-level decision making, we use the \textit{PyTrees} library \cite{pytrees}. The tree that we construct contains actions for both the switching mechanism and the functionality to take specific actions. For training our autonomous agents, we use the PyTorch framework. We use this to decide which action should be taken based on the observation.

The \textit{Cyber-Firefighter} environment is set up as an Erdos-Renyi (ER) graph~\cite{erdHos1960evolution} with $10$ nodes and a probability of $0.2$ for edge connection with a constraint that the graph must be connected, otherwise the ``fire" could not spread to all nodes. For each simulation a new ER graph is constructed with a fixed fire source, $v_f = 0$, and a fixed maximum burning time, $T_{burn}^{max}= 4$ timesteps. The simulator we use is CybORG \cite{cyborg_acd_2021}, the CAGE Challenge Scenario 2 \cite{cage_challenge_2}. The simulator records an observation state, stored as a vector of length $52$, containing $4$ bits for each of the $13$ hosts. Two of these bits encode the type of program being executed on a host and the other two represent the degree to which the host has been compromised. The full action space of cyber actions contains $145$ potential actions, however this was reduced to only $52$ meaningful actions in our approach. These actions only focus on the operational server, the defender, the enterprise servers $0-2$, and user hosts $1-4$. For each of these servers and hosts, an analyze, remove, or restore action can be performed, in addition to a limited selection of decoys specific to each host. The EBT chooses a cyber action, records it in the blackboard, and the action is then received in CybORG for execution. The blackboard variables are initially assigned from the CybORG setup and then manipulated throughout the execution of the behavior tree. These variables contain information of the state, action, and reward. During execution, the EBT also relies on action functions to perform the low-level actions. Upon completion of behavior tree execution during a single step, the blackboard variables are then written back to the CybORG variables to update the environment.

\subsection{Genetic Programming Setup}
\begin{figure}
    \centering
    \includegraphics[width=0.5\linewidth]{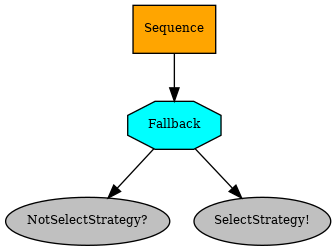}
    \caption{Baseline BT considering only strategy switching}
    \label{fig:baseline_bt}
\end{figure}

For evaluation of learning the structure of the EBT, we consider a comparison of multiple BT structures and their respective performance in the abstract environment. The GP algorithm is initialized with $16$ BTs and a baseline BT in Fig.~\ref{fig:baseline_bt}. The baseline BT is \textit{boosted} through the inclusion in each generation and preference towards crossover and mutation. In each generation, a BT is represented using a hierarchical list of strings representation, which can then be directly converted to an executable BT for evaluation. The string representation allows for simple operations to update and add new behaviors before execution. The algorithm is executed for $5$ trials of $200$ generations with approximately $5000$ simulation runs (episodes) recorded with the best fitness for each generation stored. We compare the fitness of the GP BTs, the baseline BT, and an expert manually crafted BT.

\begin{figure*}
    \centering
    \begin{subfigure}{0.32\linewidth}
        \centering
        \includegraphics[width=1.0\textwidth]{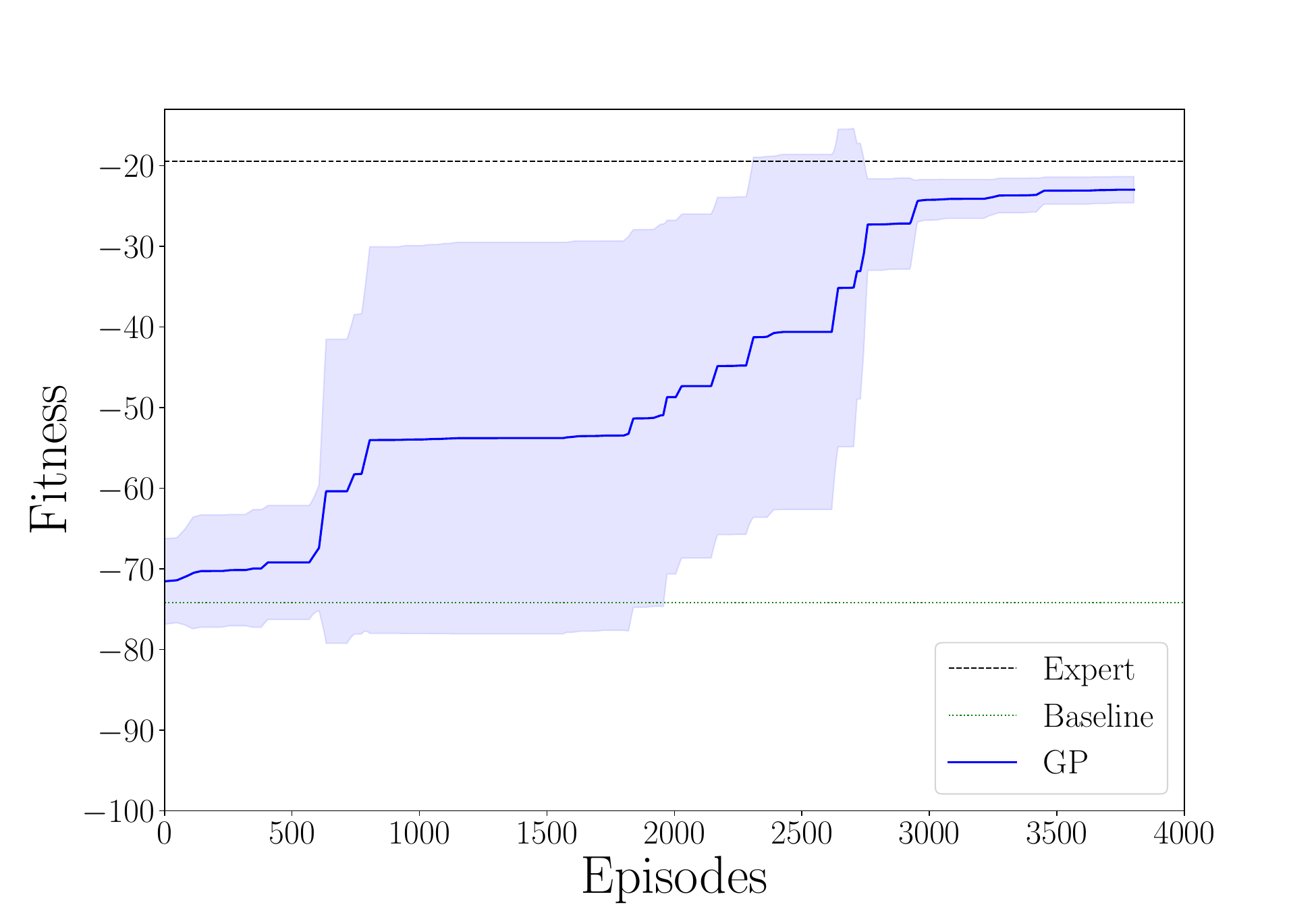}
        \caption{GP Fitness}
        \label{fig: gp_success}
    \end{subfigure} %
    \centering
    \begin{subfigure}{0.32\linewidth}
        \centering
        \includegraphics[width=1.0\textwidth]{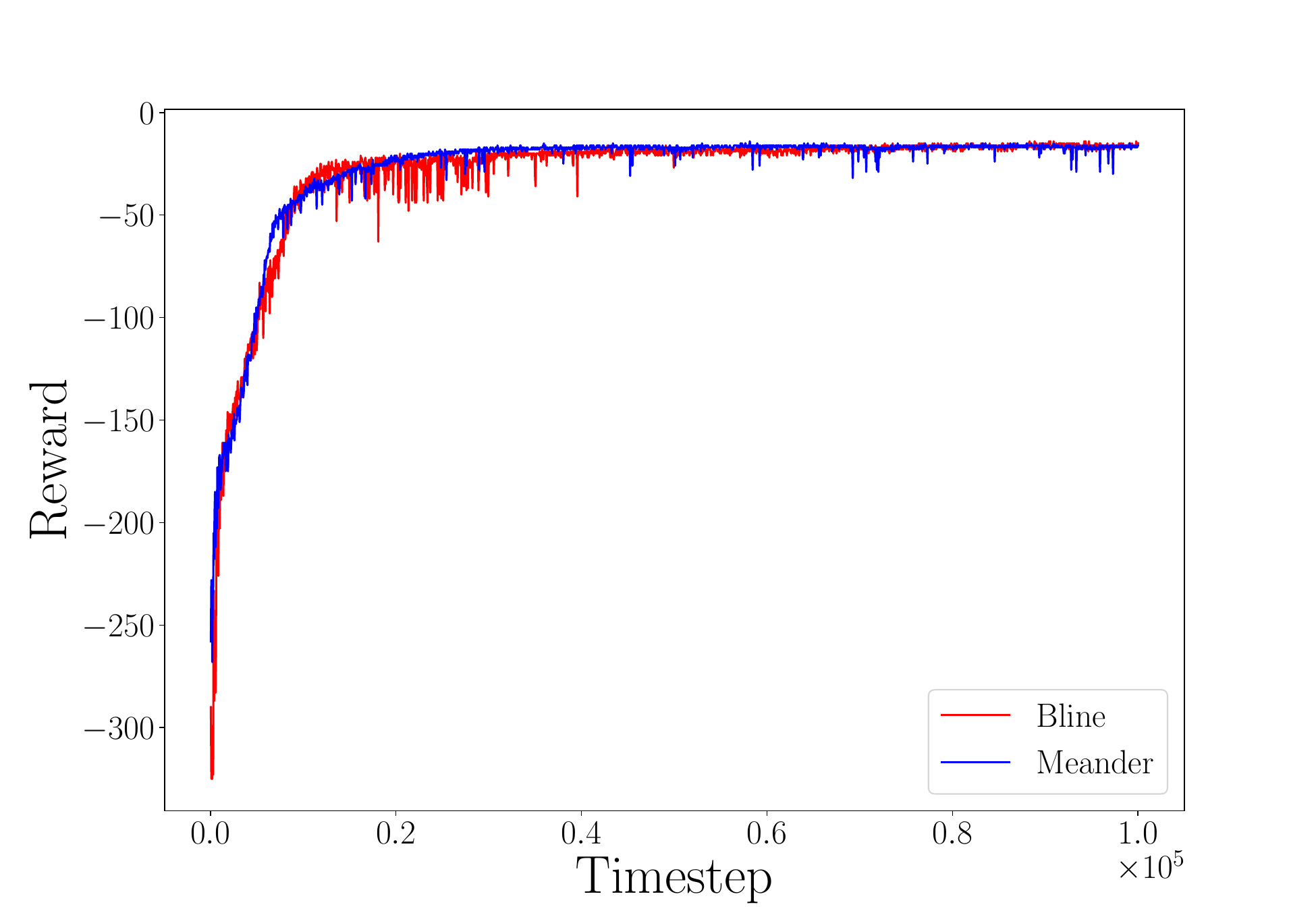}
        \caption{Cyber-Agent Controller Rewards}
        \label{fig: controller_rew}
    \end{subfigure} %
    \centering
    \begin{subfigure}{0.32\linewidth}
        \centering
        \includegraphics[width=1.0\textwidth]{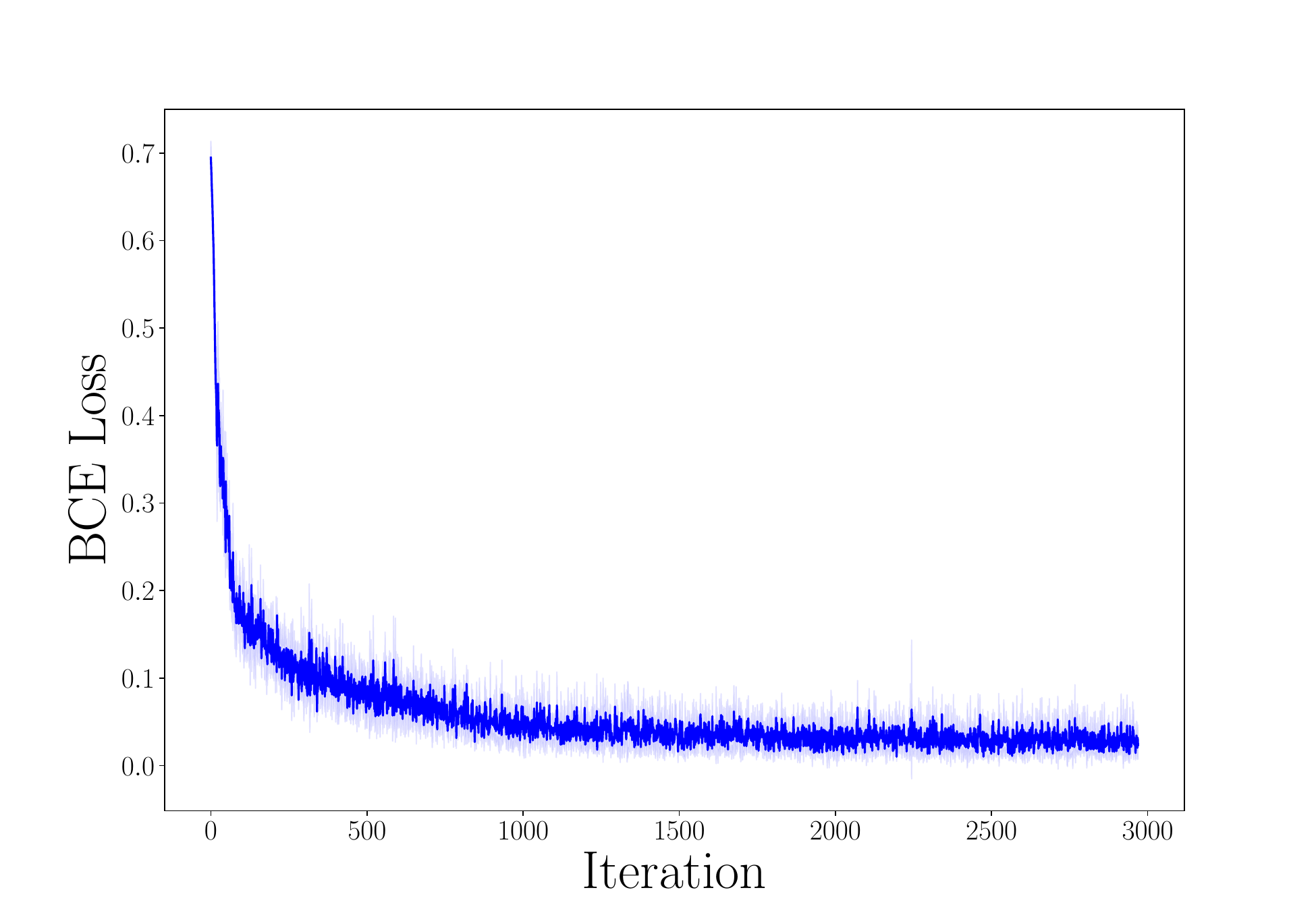}
        \caption{Strategy-Switch Policy Loss}
        \label{fig:lstm_training}
    \end{subfigure} 
    \caption{Training evaluation for learning structure and optimizing LECs. The shaded region in (a) and (c) represents the standard deviation over 5 training runs.}
    \label{fig: training_eval}
\end{figure*}

\subsection{GP Performance}
Fig.~\ref{fig: gp_success} shows the GP fitness over $5$ training runs of $4000$ episodes for the Expert, Baseline, and GPBT. The GPBT fitness (plot in blue) improves from the baseline (plot in green) up to the Expert BT (plot in black). The learned BT structure, computed from the trials, achieves the best fitness of $-19.45$ (shown in Fig.~\ref{fig:bt_structure}). This is the same fitness as the Expert BT. Therefore, we can conclude that we can
learn the structure of a BT to achieve an abstract cyber-security task with  performance similar to a BT constructed with expert knowledge.

\subsection{Optimizing Behaviors Setup}

Opposed to the standard scenario, in which the red agent selects one strategy for the entirety of an episode, our scenario sees the \textit{RedSwitch} agent that changes from a \textit{Meander} to a \textit{BLine} strategy. This strategy switching is set to randomly occur between timesteps $10-30$, simulating how an attacker may change their strategy based on information of the network structure that they can quickly obtain. The defender (EBT) interacts with the environment in the architecture in Fig.~\ref{fig:software_arch} through the \textit{blackboard} to determine and adapt from a red strategy change.

Two cyber-agent controller policies were trained based on the \textit{Meander} and \textit{BLine} strategies using a PPO policy. The policies consist of actor-critic neural networks with the actor and critic networks containing $1$ layer of size $64$.  Both of these policies were trained for $100$K observations with $1000$ episodes each in a non-deterministic manner. This was done using a multi-layered perceptron developed by the \textit{CardiffUni} team. Training was done in order to minimize the negative reward received by the blue agent from allowing the red agent to take certain actions and reach various hosts. The blue agent receives a reward of $-0.1$ per turn for each user and operational host (excluding user host $0$) the red agent has access to. A reward of $-1$ per turn is received for each enterprise and operational server the red agent has access to, or if a blue agent performs a restore operation. If the red agent performs an impact action on the operational server, the blue agent receives a reward of $-10$ per turn. Since these rewards are negative, an optimal solution should be closer to $0$, indicating better success at mitigating a red agent attack. Training rewards over $100$K timesteps for the \textit{BLine} and \textit{Meander} cyber-agent controller policies are shown in Fig.~\ref{fig: controller_rew}. The reward is maximized, asymptotically approaching $0$, indicating the training was successful. Therefore, the policies are prepared for deployment evaluation.

An LSTM model was trained in order to carry out the strategy switching. This model was trained successfully with a window size of $5$ over $5$ epochs ($3000$ optimization iterations) with $100$k observations ($1000$ episodes) using supervised Binary Cross Entropy (BCE) loss. Observations are     collected using an oracle that correctly labels each strategy, \textit{Meander} or \textit{BLine}. The supervised loss is used to optimize the predicted label of the strategy from the LSTM with the correct labelling from the oracle using a window of observations. The LSTM model consists of a neural network with $2$ layers of size $100$. The learning rate used was $1e-3$. BCE training loss for $3000$ iterations over $5$ training runs is shown in Fig.~\ref{fig:lstm_training}. A supervised approach was taken in which the model was aware when the red agent switched strategies in order to maximize the efficacy of the training. Similar to training the PPO models, these agents were trained to minimize the reward received.


We first evaluate that our cyber-agent controller policies for \textit{Meander} and \textit{BLine} integrated into the GPBT structure. The original CardiffUni solution is compared against our EBT without considering strategy switching. We evaluate using the cumulative reward over episodes for $1000$ episodes for both \textit{Meander} and \textit{BLine}. Then, we evaluate multiple EBT solutions in our BT structure (GPBT) in order to compare their performance in our scenario. Every solution begins by scanning the network over the first three timesteps to determine the initial red agent strategy. 
\begin{itemize}
    \item \textbf{CardiffUni}: The base solution with no strategy switching mechanism; once it determines the initial strategy (Me- ander), it retains it for the duration of the episode.
    \item \textbf{OracleSwitch}: Operates as an oracle that gets informa- tion directly from the simulation on the exact timestep the RedSwitch agent changes strategies during an episode.
    \item \textbf{LearnedSwitch}: (Ours) Uses the trained LSTM model to determine if the red agent has switched strategies based on its current behavior. LearnedSwitch uses OracleSwitch during training.
\end{itemize}

\subsection{Attacker Strategy Robustness}

\begin{figure}
    \centering
    \begin{subfigure}{0.49\linewidth}
        \centering
        \includegraphics[width=1.0\textwidth]{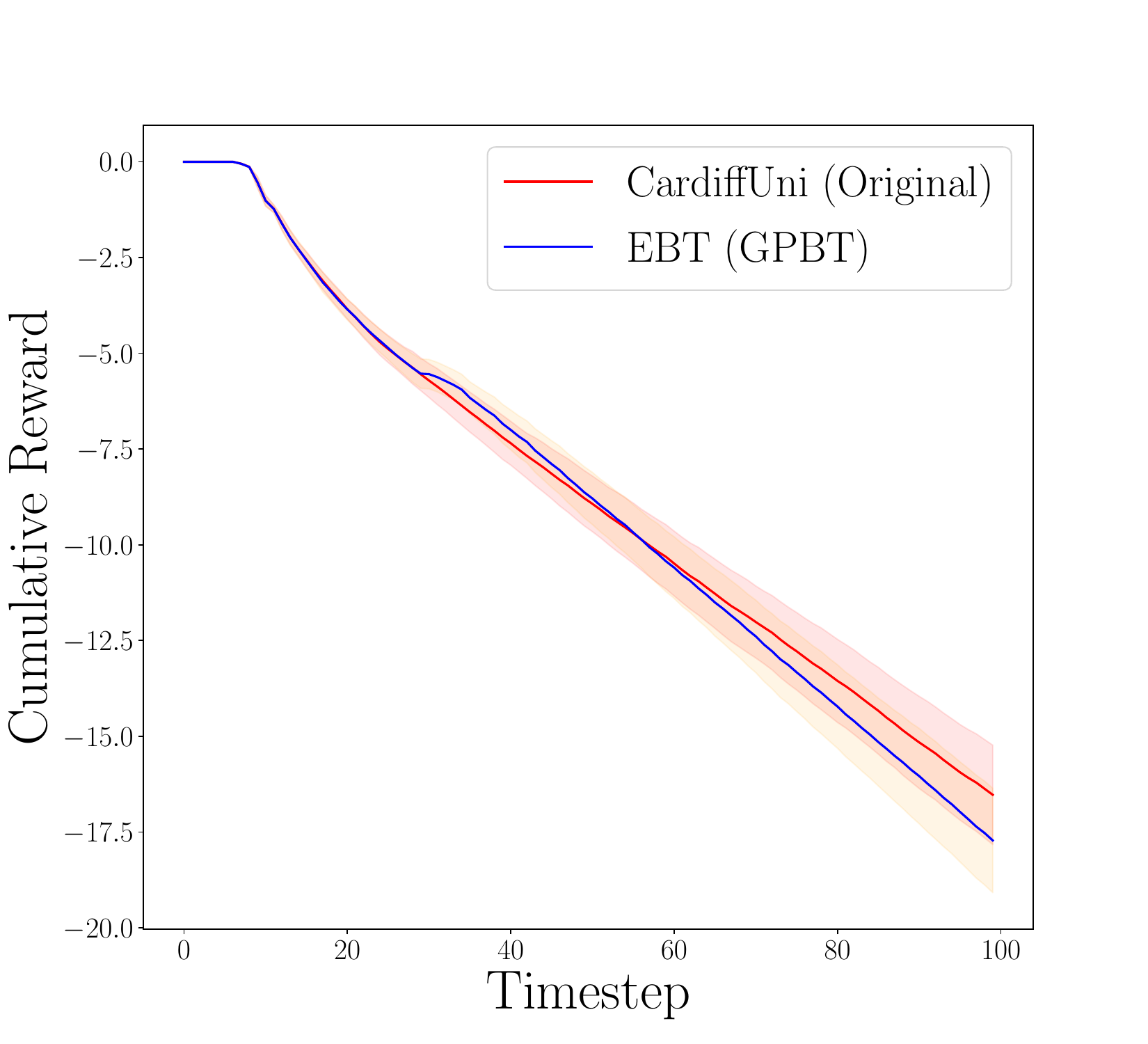}
        \caption{Meander}
        \label{fig: meander_comparison}
    \end{subfigure} %
    \centering
    \begin{subfigure}{0.49\linewidth}
        \centering
        \includegraphics[width=1.0\textwidth]{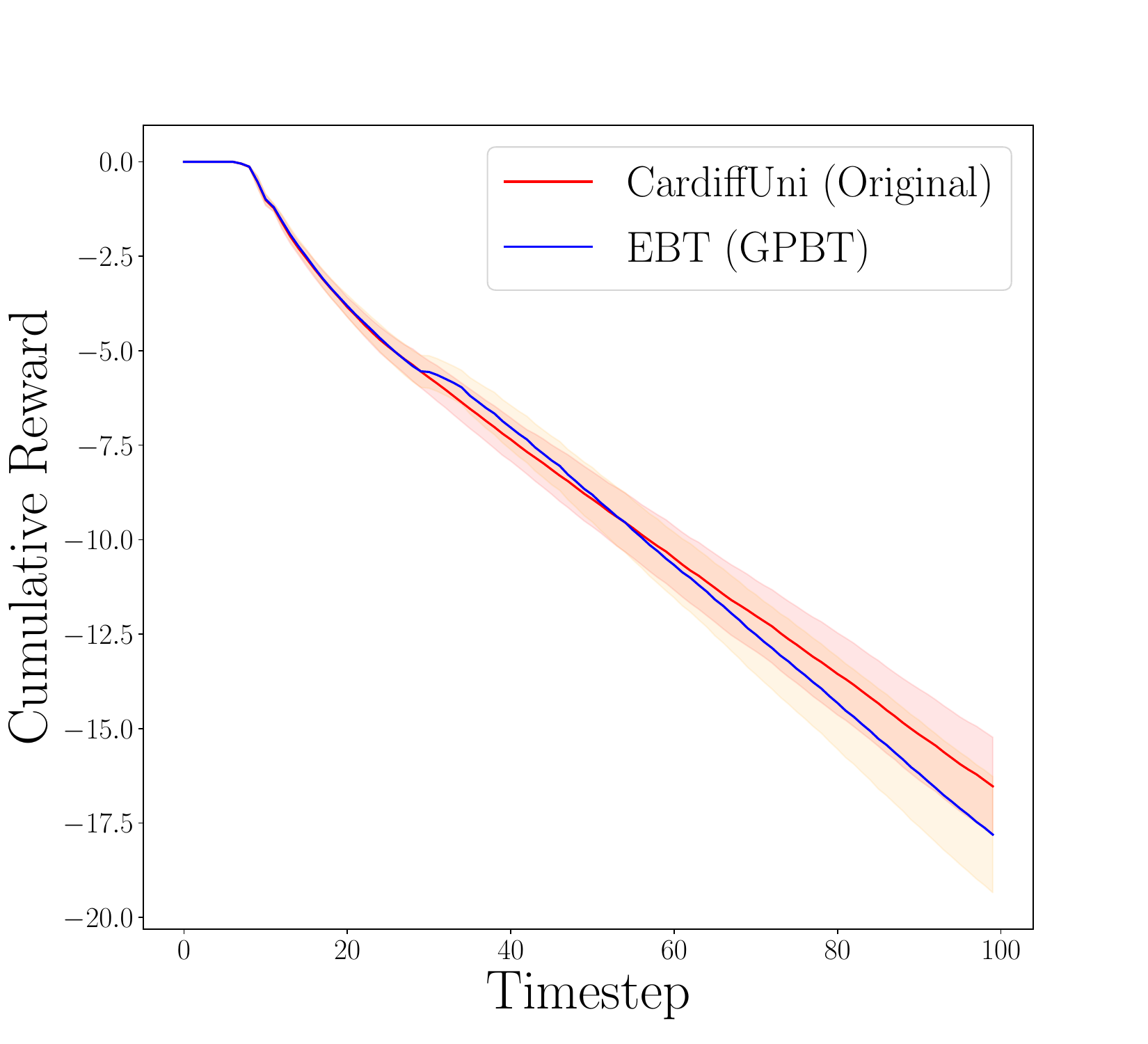}
        \caption{BLine}
        \label{fig: bline_comparison}
    \end{subfigure} \\
    \begin{subfigure}{1.0\linewidth}
        \centering
        \includegraphics[width=0.5\textwidth]{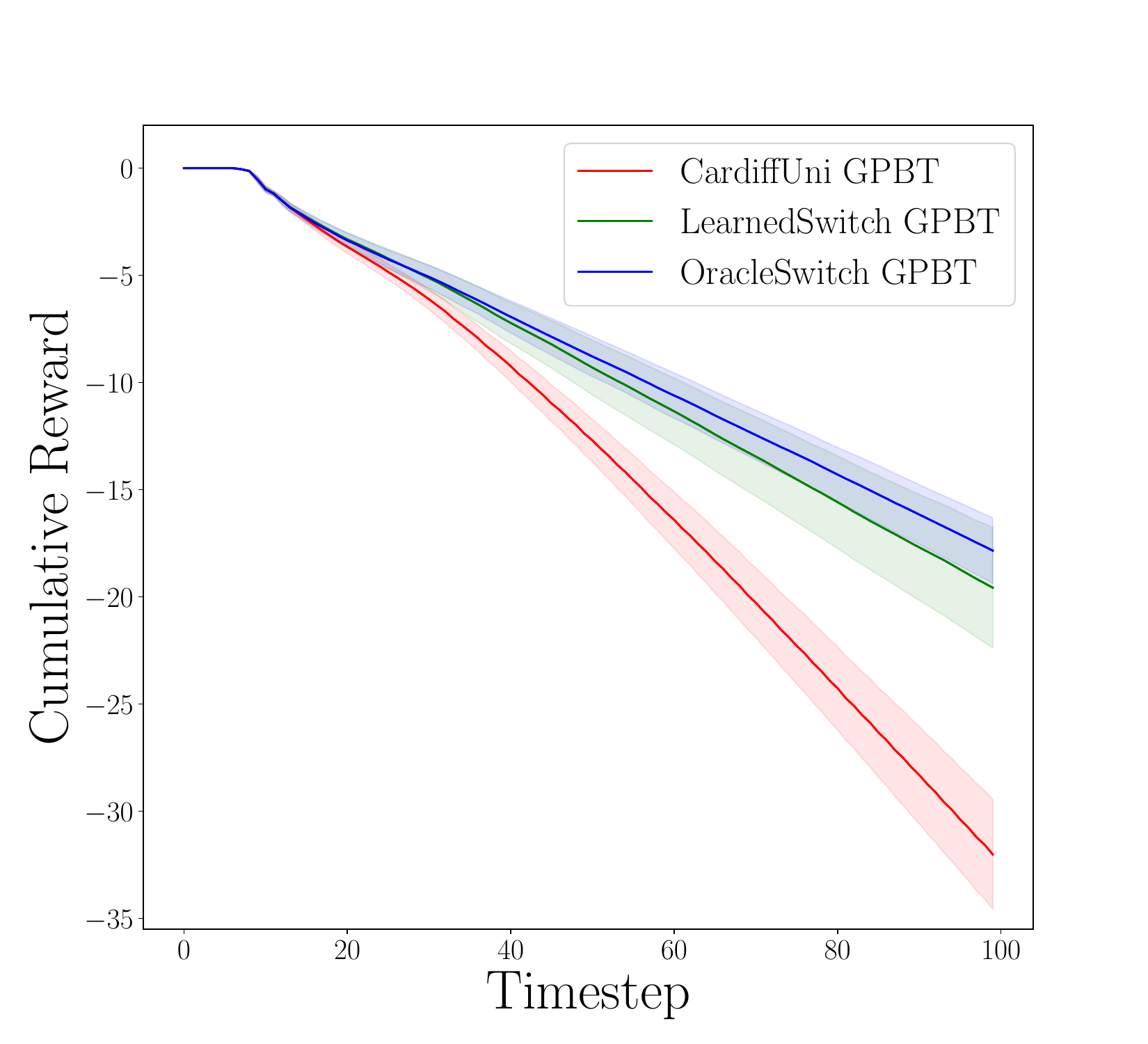}
        \caption{Strategy Switching}
        \label{fig:strat_switch_results}
    \end{subfigure}
    \caption{Cumulative Reward over time, recorded for $1000$ episodes. In (a) and (b), the EBT with LECs is compared to original approach with LECs. In (c) varying degrees of strategy switching are compared. Mean and Standard Deviation are plotted.}
    \label{fig:reward_results}
\end{figure}



In Figs.~\ref{fig: meander_comparison} and \ref{fig: bline_comparison}, the cumulative reward over is plotted for CardiffUni and the EBT, recorded for $1000$ episodes. The EBT (blue) achieves consistent results with the original CardiffUni approach (red). The EBT (blue) performs slightly below CardiffUni (red) on average over $1000$ episodes. This is likely attributed to minimal overhead by BT execution. Therefore, integrating LECs into the GPBT structure to develop an EBT still achieves consistent performance with the state-of-the-art CardiffUni approach. 

Fig.~\ref{fig:strat_switch_results}, displays the cumulative reward over time recorded for $1000$ episodes, comparing CardiffUni, OracleSwitch, and LearnedSwitch GPBTs. We found that our LearnedSwitch solution (green) shows a significant improvement of around $39\%$ over the CardiffUni solution (red). The CardiffUni solution managed an average cumulative reward of $-32.03$, while our LearnedSwitch solution achieved an average cumulative reward of $-19.56$. This indicates that a strategy switch for the cyber agent is imperative to properly defend against a red agent that can switch strategies; solely detecting its strategy at the beginning of an episode is not sufficient. It should be noted that this solution did not see a noticeable drop in performance in the range of timesteps in which the red agent could change strategies. This highlights our solution as being resilient to a sudden change in strategy, as it can rapidly detect and adapt to it. Additionally, the OracleSwitch solution (blue) resulted in an average cumulative reward of $-17.84$. This slightly outperforms the LearnedSwitch solution, which is expected due to its immediate knowledge of a strategy switch. However, the reward between the two is comparable.

\subsection{Explainability}



The CardiffUni solution used a simple sequential approach in carrying out the actions during each episode. However, we employed an EBT in our solutions, as we believe that it improves evaluation. The EBT precisely explains the operations of the simulation during each episode and acts as a runtime monitor for high-level behaviors. In particular, it allows us to visually determine when certain events are occurring, such as adapting to a new strategy or deploying a decoy. Visualization is achieved through the built-in functionality of \textit{PyTrees} to render BT execution~\cite{pytrees}. This provides us with an interface for following the execution of the components throughout the runtime of the EBT. Also, the EBT simplifies the task of performing the strategy switch during an episode. It includes designated nodes to evaluate the current red agent strategy and then switch its own strategy, if needed. These actions are explicitly independent as illustrated in the EBT, both from each other and from other high-level actions. This overall assists in improving control flow of the program.
\section{Conclusions \& Future Work}
\noindent
In this work, we describe an approach for designing long-term autonomous cyber-defense agents enabled by Evolving Behavior Trees (EBTs). We utilize a GP algorithm and develop a novel abstract cyber environment to learn the high-level structure of the EBT. Then, we optimize the EBT in a realistic cyber environment with LECs for reactive strategy switching to mitigate complex and dynamic cyber-attacks. The structure of the EBT is modular and generalizable to autonomous cyber-defense in a network. The performance of learning the structure is evaluated in the abstract cyber environment where we demonstrate how the structure learned will promote visibility and mitigate an attack. We then evaluate the integration of LECs in the EBT in the CybORG simulation environment using CAGE Challenge Scenario 2. Furthermore, we develop a software architecture to integrate the EBT with CybORG. Our results demonstrate that our proposed approach against the attacker strategy switching shows a $39\%$ improvement in average reward compared to a state-of-the-art approach, and provides explainability for use in runtime monitoring of key events.

Our future works will focus on scalability, emulation, and expansion to new attack scenarios. We aim to expand our set of available defense strategies to cover a multitude of attacks. 
We will also research methods to train individual policies that can generalize to a subset of red agent strategies. Additionally, we seek to evaluate our approach in an emulation environment to demonstrate the ability to deploy our solution in real network systems.

\bibliographystyle{IEEEtran}
\bibliography{ref}

\end{document}